\documentclass[letterpaper, 10 pt, conference]{ieeeconf}

\usepackage{multirow}
\usepackage{multicol}
\usepackage{graphicx}
\usepackage{xspace}
\usepackage[table,x11names,svgnames,dvipsnames]{xcolor}
\usepackage{caption}
\usepackage{wrapfig}
\usepackage{bbding}  
\usepackage{pifont}  
\usepackage[ruled,linesnumbered]{algorithm2e}
\newcommand{\nonl}{\renewcommand{\nl}{\let\nl\oldnl}}
\let\oldnl\nl

\usepackage{booktabs}
\usepackage{float}
\usepackage{amsmath}
\usepackage{amssymb}
\usepackage{hyperref}

\newcommand{\oursfull}[0]{Selective Flow Alignment\xspace}
\newcommand{\ours}[0]{SeFA\xspace}
\newcommand{\ourmethod}[0]{Selectively Align\xspace}

\usepackage{colortbl}
\definecolor{ourcolor}{HTML}{99e0eb}
\definecolor{ourblue}{HTML}{27a2c3}

\definecolor{tablecolor}{HTML}{ccf2f5} 

\definecolor{tablecolor2}{HTML}{ffcdb4}
\definecolor{citecolor}{HTML}{fe7b5b}
\definecolor{grey}{rgb}{0.9, 0.9, 0.9}
\usepackage{amssymb}

\usepackage{listings}
\definecolor{gred}{rgb}{0.859,0.267,0.216}
\definecolor{ggreen}{rgb}{0.059,0.616,0.345}

\definecolor{deepblue}{HTML}{27a2c3}

\definecolor{deepred}{HTML}{fe7b5b}

\newcommand{\dd}[2]{$#1\scriptstyle{\pm#2}$}
\newcommand{\ddbf}[2]{\cellcolor{tablecolor}$\mathbf{#1\scriptstyle{\pm#2}}$}

\newcommand{\cc}[1]{$#1$}
\newcommand{\ccbf}[1]{\cellcolor{tablecolor}$\mathbf{#1}$}

\usepackage{subfigure}
\usepackage{stfloats}

\title{\LARGE \bf SeFA-Policy: Fast and Accurate Visuomotor Policy Learning \\ with Selective Flow Alignment}

\author{\authorblockN{Rong Xue\authorrefmark{2}\authorrefmark{1},
Jiageng Mao\authorrefmark{2}\authorrefmark{1},
Mingtong Zhang\authorrefmark{2},
Yue Wang\authorrefmark{2}}
\authorblockA{\authorrefmark{2}University of Southern California \authorrefmark{1}Equal Contribution}
\authorblockA{\href{https://rongxuezoe.github.io/SeFAPolicy-homepage/}{\color{deepblue}\textbf{SeFA-Policy}\xspace}}
}

\begin{document}

\twocolumn[{%
\renewcommand\twocolumn[1][]{#1}%
\maketitle
\begin{center}
    \captionsetup{type=figure}
\includegraphics[width=1.\linewidth]{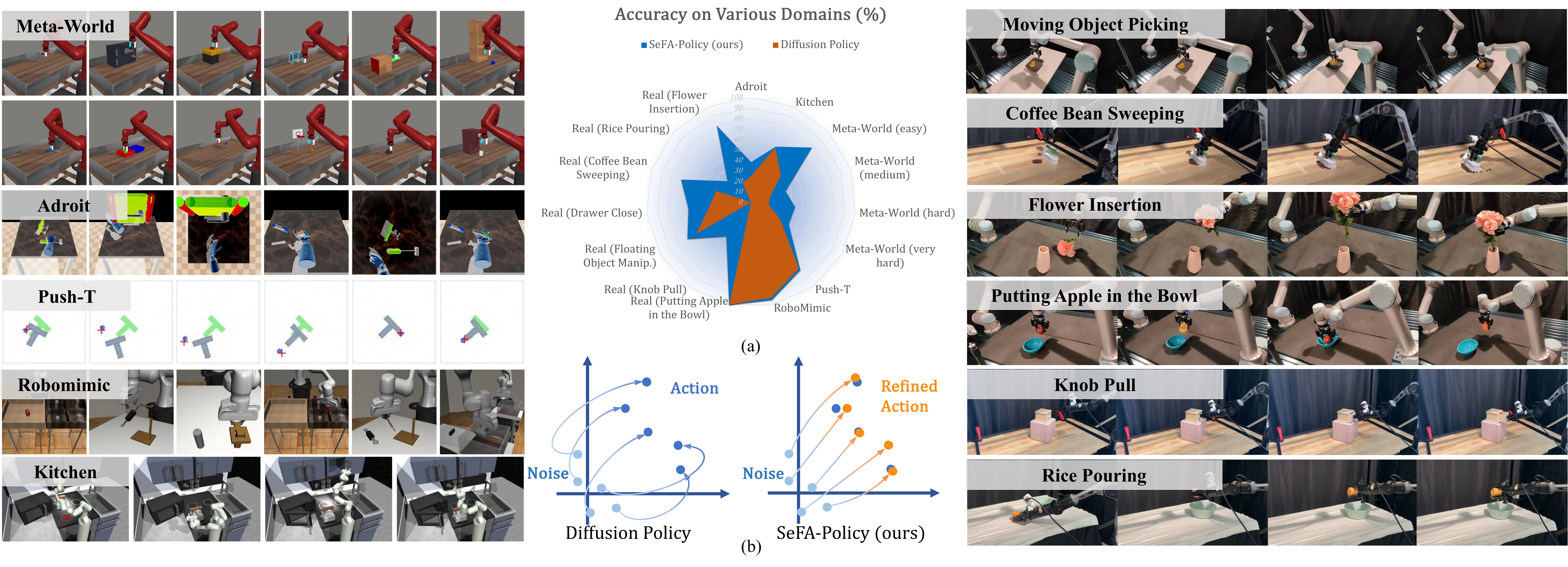}
\caption{\textbf{\oursfull} (\textbf{\ours}) is a visual imitation learning algorithm that utilizes rectified flow with selective alignment, achieving superior effectiveness in diverse simulation and real-world tasks, with a significant inference acceleration. (a) Accuracy on various domains. (b) Sampling flow from noise to action of \ours and Diffusion Policy.}
\label{fig:teaser}
\end{center}
}]

\begin{abstract}

Developing efficient and accurate visuomotor policies poses a central challenge in robotic imitation learning. While recent rectified flow approaches have advanced visuomotor policy learning, they suffer from a key limitation: After iterative distillation, generated actions may deviate from the ground-truth actions corresponding to the current visual observation, leading to accumulated error as the reflow process repeats and unstable task execution. We present \oursfull (\ours), an efficient and accurate visuomotor policy learning framework. \ours resolves this challenge by a selective flow alignment strategy, which leverages expert demonstrations to selectively correct generated actions and restore consistency with observations, while preserving multimodality. This design introduces a consistency correction mechanism that ensures generated actions remain observation-aligned without sacrificing the efficiency of one-step flow inference. Extensive experiments across both simulated and real-world manipulation tasks show that \ours surpasses state-of-the-art diffusion-based and flow-based policies, achieving superior accuracy and robustness while reducing inference latency by over 98\%. By unifying rectified flow efficiency with observation-consistent action generation, \ours provides a scalable and dependable solution for real-time visuomotor policy learning. 
Code is available on \href{https://github.com/RongXueZoe/SeFA}{\color{deepblue}\textbf{\ours code}\xspace}. 

\end{abstract}


\section{Introduction}

Imitation learning relies on accurate action predictions and fast inference to successfully perform complicated real-world tasks.
Generative modeling techniques such as Diffusion Policy~\cite{chi2023diffusion} have recently achieved strong performance in complex manipulation tasks, but their reliance on multi-step iterative denoising makes them computationally expensive and unsuitable for real-time control. Flow-based models have emerged as a promising alternative, enabling fewer-step action generation~\cite{hu2024adaflow} and rectification~\cite{liu2023flow} by transporting nearly straight from noise to action space, thus significantly reducing inference latency.

Despite the efficiency, few-step sampling introduces discretization error, and rectification introduces inconsistency between observation and action during distillation. 
When applying rectification, the reflow policy~\cite{liu2023flow} is trained upon the noise-action pairs generated by a well-trained policy. However, the generated actions are not the same as ground-truth actions, i.e., the generated action and visual observation pair could not exactly match. In terms of diffusion-based models, the predicted action is distinct from the actions reflected in the visual condition.
Such inconsistency might be tolerable in image generation where perceptual similarity suffices, but in robotic control, even minor inconsistencies between observations and actions can accumulate and lead to task failures. This distillation-induced inconsistency therefore represents a fundamental barrier to deploying flow-based policies in effective real-time visuomotor control.

To overcome this limitation, we propose \textbf{\oursfull} (\textbf{\ours}), a flow-based visuomotor policy with a selective alignment strategy. The straight paths in flow-based models are computationally efficient because they can be sampled in a few or even one step. However, the straightening process~\cite{liu2023flow} often accumulates errors from the base model, which leads to the observation–action inconsistency. \ours leverages expert demonstrations to align the sampling paths with observations while maintaining the straightness of the paths.
Crucially, this alignment is applied in a selective manner, preserving action diversity and multimodality while eliminating harmful mismatches. 
By combining efficiency in straight sampling paths with observation-consistent alignment, \ours enables one-step action synthesis that is both fast and reliable for real-time visuomotor control.

We summarize our main contributions as follows:
\begin{itemize}
\item \oursfull (\ours): We introduce a novel visuomotor policy learning framework that improves flow-based models with a selective alignment strategy to address the observation–action inconsistency problem.

\item Consistency Correction with Efficiency: Our method achieves one-step action generation that is both observation-aligned and computationally efficient, preserving efficiency while ensuring robustness in control.

\item Extensive Evaluation: We demonstrate \ours’s effectiveness across diverse simulated and real-world robotic tasks, where it consistently surpasses diffusion-based and flow-based baselines in accuracy, robustness, and inference speed.
\end{itemize}

\begin{figure*}[!t]
\begin{center}
\centering
\captionsetup{type=figure}

\includegraphics[width=1\linewidth]{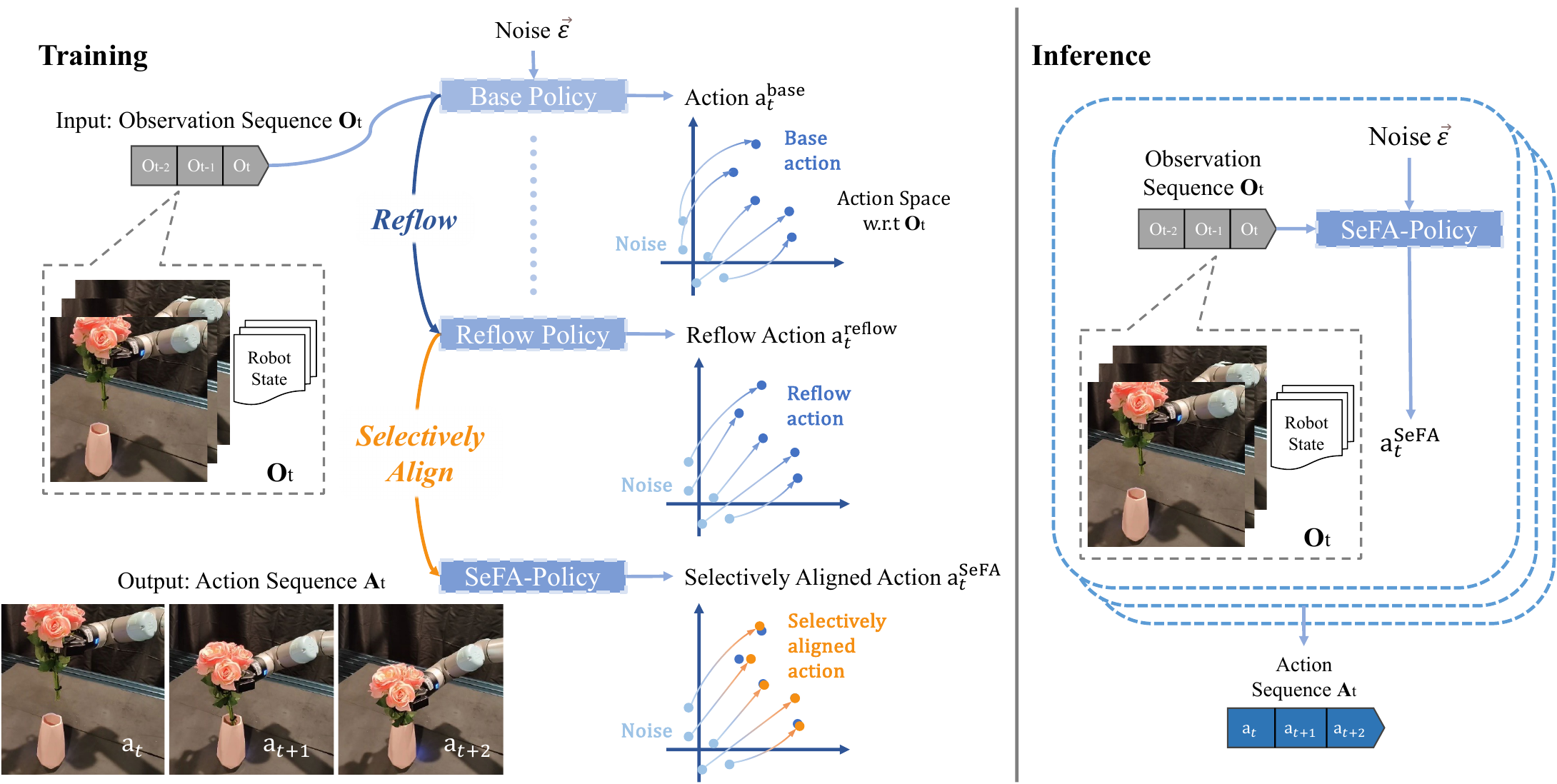}
\vspace{-0.6cm}
\caption{\textbf{Overview of \ours.} We train a visuomotor policy in an iterative manner to transport straight between noise distribution and target action space, hence enabling lightning one-step sampling during inference. The action flow is selectively \textit{aligned} with observations, lowering the potential accumulated error brought by multiple reflows.}

\vspace{-0.8cm}
\label{fig:overview}
\end{center}
\end{figure*}
\section{Related Work}
\noindent\textbf{Diffusion Models in Robotics.}
Diffusion models excel at representing complex multimodal distributions with stable training dynamics and hyperparameter robustness, gaining traction across robotics applications including motion planning \cite{jannerPlanningDiffusionFlexible2022a, luo2024potential, carvalho2023mpd, 10610519, huang2023diffusion}, imitation learning \cite{pearceImitatingHumanBehaviour2022, chiDiffusionPolicyVisuomotor2023b, pmlr-v229-ha23a, pmlr-v229-xian23a, liCrosswayDiffusionImproving2024, Ze2024DP3, Wang-RSS-24, Chen-RSS-24, sridhar2023memoryconsistent, zhao2024aloha, Chi-RSS-24}, goal-conditioned imitation learning \cite{reussGoalConditionedImitationLearning2023a, Reuss-RSS-24, chen2023playfusion, language-control-diffusion}, and grasp prediction \cite{urain2022se3dif}. However, their iterative denoising process renders them impractical for real-time robotics applications. Our proposed \ours overcomes this limitation through rectified action flow, enabling deterministic, fast action synthesis without compromising accuracy.

Recent work has explored flow matching~\cite{lipmanflow}, a diffusion variant, for representing complex continuous action distributions. 
AdaFlow~\cite{hu2024adaflow} introduces a variance-adaptive ODE solver with adjustable step sizes, but only achieves one-step generation for uni-modal distributions due to lacking reflow, and hasn't been validated on real robots or comprehensive simulation domains. 
$\pi_0$~\cite{pi0} leverages pre-trained Vision-Language Models with flow matching for high-frequency action generation, but still requires 10 integration steps. In contrast, our method maintains high precision even with one-step prediction.

\noindent\textbf{Accelerating Diffusion Models for Robotics.}
Efforts to speed up diffusion models have been explored extensively in both image generation \cite{karrasElucidatingDesignSpace2022a, songConsistencyModels2023a, songDenoisingDiffusionImplicit2020, kim2024consistency} and robotics.
However, these methods often require complex distillation processes or introduce constraints, such as overly smooth trajectories in Dynamical Motion Primitives (DMPs) \cite{Scheikl2024MPD}. Streaming Diffusion Policy (SDP)~\cite{høeg2024streamingdiffusionpolicyfast} and related approaches like Rolling Diffusion~\cite{pmlr-v235-ruhe24a} and Temporally Entangled Diffusion~\cite{zhangTEDiTemporallyEntangledDiffusion2023} improve speed through parallelization or buffering, but they often incur significant memory overhead or require intricate implementation.

Overall, while significant progress has been made in accelerating diffusion models for robotics, achieving real-time performance without sacrificing accuracy remains a challenge. Our work builds on these advancements by introducing \ours, which leverages rectified flow to achieve fast and accurate visuomotor control and introduces a selective \textit{alignment} strategy to mitigate potential accumulated errors from multiple reflows. By replacing the iterative denoising process with a deterministic coupling, \ours offers a streamlined and efficient solution for real-time robotic applications.

\section{Method}

A visuomotor policy solves the task of observing a sequence of visual observations and predicts the next action to execute in the environment. We formulate \ours-Policy as a flow-based model in \S\ref{sec:modeling}.
Then, we introduce our proposed \textit{\oursfull} strategy to alleviate performance degradation and improve the performance in \S\ref{sec:refine}. 

\subsection{Flow-based Model}\label{sec:modeling}
We begin by formalizing our approach to action space modeling. The fundamental objective is to establish a bijective mapping between standard Gaussian noise $\mathbf{a}_T \in \mathbb{R}^d$ and the target action distribution $\mathbf{a}_0 \in \mathbb{R}^d$ given the visual observations $\mathbf{O}$. During the initial training phase, we define a drift vector field $v: \mathbb{R}^d \rightarrow \mathbb{R}^d$ that guides the flow along trajectories approximating the direct linear path from $\mathbf{a}_T$ to $\mathbf{a}_0$. This is achieved by minimizing the expected squared deviation between the drift and the ideal direction $(\mathbf{a}_0 - \mathbf{a}_T)$ through the following least squares optimization problem:
\begin{equation}\label{equ:drift}
\min_{v} \int_0^1\mathbb{E} \left[ \vert\vert{( \mathbf{a}_0 - \mathbf{a}_T) - v\big (\mathbf{a}_t, t, \mathbf{O}\big)} \vert\vert^2\right] \text{d}t, 
\end{equation}
where $\mathbf{a}_t$ is the linear interpolation  of  $\mathbf{a}_T$ and  $\mathbf{a}_0$, i.e., $\mathbf{a}_t = \frac{t}{T} \mathbf{a}_T + \frac{T-t}{T} \mathbf{a}_0$, where $t \in [0,T]$. In the following, we will omit the conditioning on visual observations $\mathbf{O}$ for simplicity.
Naturally, $\mathbf{a}_t$ follows the ODE of $\text{d} \mathbf{a}_t =(\mathbf{a}_0 - \mathbf{a}_T)\text{d}t$, where any update of $\mathbf{a}_t$ requires the information of the target clean action $\mathbf{a}_0$.  
By fitting the drift $v$ with $\mathbf{a}_0 - \mathbf{a}_T$, the action flow causalizes the paths of linear interpolation $\mathbf{a}_t$, relieving the burden of involving the target action (which is unknown during inference) when simulating the ODE flow.

The solution to Equation~\eqref{equ:drift} constitutes our base policy network, denoted as $v^\text{base}$. When provided with a noise sample $\mathbf{a}_T$, the base policy generates the corresponding action $\mathbf{a}_0$. 

To make a distinction, $v^\text{base}$ is trained with the random noise and the groundtruth action pairs $(\mathbf{a}_T, \mathbf{a}_0)$, and the derived coupling is denoted as $(\mathbf{a}^\text{base}_T, \mathbf{a}^\text{base}_0)$. The action flow induced between $\mathcal{N}(0, \mathbf{I})$ and $\mathcal{A}$ is:
\begin{equation}
\text{d}\mathbf{a}^\text{base}_t = v^\text{base}(\mathbf{a}^\text{base}_t,t)\text{d}t, \quad t\in[0,T],
\label{equ:vbase}
\end{equation}
which converts the noise $\mathbf{a}^\text{base}_T \in \mathcal{N}(0, \mathbf{I})$ in the coupling $(\mathbf{a}^\text{base}_T, \mathbf{a}^\text{base}_0)$ to the action $\mathbf{a}^\text{base}_0$ which follows the conditioned expert action distribution.

\textbf{Reflow.}
After the drift $v^{\text{base}}$ is estimated, we can accelerate inference by training a reflow policy~\cite{liu2023flow} that straightens the sampling paths.
In our settings, we first randomly sample noise $\mathbf{a}_T^\text{base} \sim \mathcal{N}(0, \mathbf{I})$ and then forward generate $\mathbf{a}_0^\text{base}$ following Equation~\eqref{equ:vbase}. This coupling $(\mathbf{a}^\text{base}_T, \mathbf{a}^\text{base}_0)$ ensures optimal transport efficiency—specifically, it guarantees that the transport cost remains lower than that of any arbitrary (action$\times$noise) pairing across all convex cost functions, a property that follows directly from Jensen's inequality. Through this formulation, we establish a principled deterministic mapping between the action space $\mathcal{A}$ and the standard Gaussian distribution $\mathcal{N}(0, \mathbf{I})$.

Following \cite{liu2023flow}, we train the reflow policy network using the optimal coupling $(\mathbf{a}^\text{base}_T, \mathbf{a}^\text{base}_0)$ as the substitution of the former pairs.
The reflowed action flow $v^\text{reflow}$ satisfies:
\begin{equation}
\text{d}\mathbf{a}^\text{reflow}_t = v^\text{reflow}(\mathbf{a}^\text{reflow}_t,t)\text{d}t, \quad t\in[0,T],
\end{equation}
where $\mathbf{a}^\text{reflow}_t$ is the linear interpolation of $\mathbf{a}^\text{base}_T$ and  $\mathbf{a}^\text{base}_0$, i.e., $\mathbf{a}^\text{reflow}_t = \frac{t}{T} \mathbf{a}^\text{base}_T + \frac{T-t}{T} \mathbf{a}^\text{base}_0$, where $t \in [0,T]$.
This procedure straightens the paths of the flows. The straighter the paths are, the smaller the time-discretization error in numerical simulation will be. Perfectly straight paths can be exactly simulated with a single Euler step. This addresses the very bottleneck of high inference cost in existing continuous-time ODE-based models, such as the Diffusion Policy~\cite{chi2023diffusion} built upon Probability Flow ODE~\cite{song2021scoresde}.

\begin{figure*}[!t]
    \centering
    \includegraphics[width=.8\textwidth]{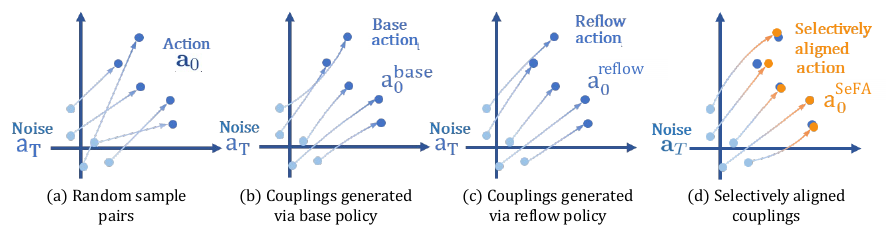}
    \vspace{0.2cm}
    \caption{\textbf{Sampling trajectories of \ours-Policy at different stages.} Randomly sampled pairs in (a) have crossing flows. \textit{Coupling}s in (b) have been rewired so they do not intersect with each other at the same denoising timestep.  The trajectories in (c) and (d) are nearly straight. }
    \vspace{-0.7cm}
    \label{fig:flow}
\end{figure*}

\subsection{Selective Flow Alignment}\label{sec:refine}

It is noteworthy that $v^\text{reflow}$ is trained upon the couplings generated by $v^\text{base}$. However, the generated actions $\mathbf{a}^\text{reflow}_{0}$ are not always the same as ground-truth actions, i.e., the generated action and visual observation pair could not exactly match. In terms of diffusion-based models, the generated action could be distinct from the actions reflected in the visual condition. 
Therefore, although reflow process can accelerate inference, it introduces inconsistency between observation and action. Unlike image generation where minor output variations are acceptable or sometimes even desired, even slight discrepancies in the generated actions can lead to task failures in robot learning, and the error could accumulate as the reflow procedure repeats.

To address this performance degradation introduced by reflow, we introduce a selective flow alignment strategy. For each generated action $\mathbf{a}^\text{reflow}_{0}$, we methodically search the ground-truth action dataset to identify its nearest neighbor. When the Euclidean distance between these actions falls below a predetermined threshold $\epsilon$, we replace the generated action with its ground-truth counterpart. We select $\epsilon$ heuristically to balance refinement capability and efficiency: too small values limit available expert actions for refinement, while excessively large values increase computational cost without improving selection quality. Since action spaces are conditioning-dependent, this approach maintains fixed conditioning while identifying the optimal coupling. This strategy effectively addresses two critical scenarios: (1) When a suitable nearest neighbor exists within the threshold, the prediction error attributable to the reflow process is rectified, thereby preserving the least transport costs property; and (2) When the reflow model generates a viable alternative solution that significantly differs from the expert policy, we intentionally preserve this inherent multimodality to maintain the richness of the action distribution.

After the selective alignment, we obtain a new coupling $(\mathbf{a}^\text{SeFA}_T, \mathbf{a}^\text{SeFA}_0)$. This coupling should maintain the property of optimal transport efficiency, while the actions $\mathbf{a}^\text{SeFA}_0$ are aligned with expert demonstrations. We can train the \ours-Policy network using $(\mathbf{a}^\text{SeFA}_T, \mathbf{a}^\text{SeFA}_0)$:
\begin{equation}
\text{d}\mathbf{a}^\text{SeFA}_t = v^\text{SeFA}(\mathbf{a}^\text{SeFA}_t,t)\text{d}t, \quad t\in[0,T],
\end{equation}
where $\mathbf{a}^\text{SeFA}_t$ is the linear interpolation of $\mathbf{a}^\text{SeFA}_T$ and  $\mathbf{a}^\text{SeFA}_0$, i.e., $\mathbf{a}^\text{SeFA}_t = \frac{t}{T} \mathbf{a}^\text{SeFA}_T + \frac{T-t}{T} \mathbf{a}^\text{SeFA}_0$, where $t \in [0,T]$. The drift should follow the ODE of $\text{d} \mathbf{a}^\text{SeFA}_t =(\mathbf{a}^\text{SeFA}_0 - \mathbf{a}^\text{SeFA}_T)\text{d}t$.

\section{Simulation Experiments}

We systematically evaluate \ours-Policy on 66 tasks from 5 benchmarks in simulation, including Adroit~\cite{rajeswaran2017dapg}, RoboMimic~\cite{mandlekar2021matters}, Meta-World~\cite{yu2020metaworld}, Franka Kitchen~\cite{lynch2019play}, and Push-T~\cite{florence2022ibc}.

\subsection{Algorithm}~\label{sec:algo}

We first elaborate the training procedure of reflow in Algorithm~\ref{alg:reflow} and our \textit{\oursfull} in Algorithm~\ref{alg:refine}.






\begin{algorithm}[ht]
\caption{\textit{Reflow} and \textit{coupling} generation}\label{alg:reflow}

\KwIn{Base policy with velocity estimation $v^{\text{base}}$. Number of \textit{coupling}s $N$.}
\nonl\textbf{Procedure:}

\tcp{Coupling generation}
\For{$i=1$ to $N$}{
Sample noise $\mathbf{a}_T^{\text{base}} \sim \mathcal{N}(0, \mathbf{I})$.

Generate action $\mathbf{a}_0^{\text{base}}$ following $\text{d}\mathbf{a}_t = v^{\text{base}}(\mathbf{a}_t^{\text{base}},t)\text{d}t$ starting from noise $\mathbf{a}_T^{\text{base}}$.

Construct \textit{coupling} $(\mathbf{a}_T^{\text{base}}, \mathbf{a}_0^{\text{base}})$.
}

\tcp{Training}

Initialize parameters of $v^{\text{reflow}} \doteq v^{\text{base}}$.

\While{terminal condition}{
Sample timestep $t\sim \text{Uniform}([0,1])$. \\
Compute $\mathbf{a}_t^{\text{base}}=t\mathbf{a}_T^{\text{base}} +(T-t)\mathbf{a}_0^{\text{base}}$. \\
Evaluate $\mathbb{E}\big[\vert \vert\mathbf{a}_0^{\text{base}}-\mathbf{a}_T^{\text{base}} - v^{\text{reflow}}(\mathbf{a}_t^{\text{base}}, ~t )\vert \vert^2 \big]$. \\
Update parameters of $v^{\text{reflow}}$.}

\KwOut{Reflow policy. \textit{Coupling} $(\mathbf{a}_T^{\text{reflow}}, \mathbf{a}_0^{\text{reflow}})$.}
\end{algorithm}

\begin{algorithm}[ht]
\caption{\textit{\oursfull} (\textit{\ours})}\label{alg:refine}

\KwIn{Target action $\mathbf{a}^{\text{reflow}}_0$ and corresponding visual observation sequence $\mathbf{O}$. Action distance threshold $\delta$.}
\nonl\textbf{Procedure}: 

Find the nearest ground-truth condition to $\mathbf{O}$, with its corresponding action $\mathbf{a}_0^{\text{reflow},*}$.

\eIf{$\|\mathbf{a}_0^{\text{reflow}},\mathbf{a}_0^{\text{reflow},*}\|<\delta$}{
$\mathbf{a}_0^{\text{SeFA}} \doteq \mathbf{a}_0^{\text{reflow},*}$}
{$\mathbf{a}_0^{\text{SeFA}} \doteq \mathbf{a}_0^{\text{reflow}}$}

\KwOut{{\ourmethod}ed target action $\mathbf{a}_0^{\text{SeFA}}$.}

\end{algorithm}

\subsection{Effectiveness}

\textbf{Comparison with State-of-the-Art Methods.} We benchmark \ours-Policy against two leading baselines: Diffusion Policy~\cite{Chi-RSS-23} and the flow-based AdaFlow Policy~\cite{hu2024adaflow}. To ensure a rigorous and fair comparison, all experimental settings, including training epochs, random seeds, learning rate schedules, and image resolutions, are held constant across all methods. A summary of the results is presented in Table~\ref{table:main simulation}. We deliberately avoided tuning hyperparameters for each task to maintain a fair experimental protocol. We observe that \ours-Policy achieves a success rate exceeding 80\% on 31 tasks, whereas Diffusion Policy surpasses this threshold on only 16 tasks. The average success rate of \ours-Policy attains \textbf{62.3}\%, substantially outperforming Diffusion Policy, which achieves only 38.9\%. 
The policies trained on Franka Kitchen are only given low-dimensional conditioning to test our method without visual inputs. Even under these conditions, \ours-Policy attains 100\% accuracy across all tasks, outperforming the state-of-the-art baseline and demonstrating strong robustness to the modality of conditioning. 

The superior performance of our method compared to Diffusion Policy can be attributed primarily to the nature of the sampling trajectories induced by the underlying integration schedule. Diffusion Policy often generated unstable, oscillatory motions and unnecessary deviations from the optimal path. This observation is consistent with prior findings~\cite{Chi-RSS-23}, which highlight that the integration schedule must be well-matched to the underlying data distribution. In robotic control tasks, where action distributions are typically concentrated, flow matching schedules facilitate straighter sampling trajectories and reduce integration error. By capitalizing on this property, our approach achieves greater accuracy and stability in robotic control compared to variance-preserving diffusion schedules. Compared to AdaFlow, the state-of-the-art flow-based policy, our method reduces the discretization error by introducing the \textit{\ours} strategy. Besides, AdaFlow utilizes adjustable integration steps that are typically greater than $1$, resulting in lower efficiency compared to our one-step prediction approach.

\textbf{Comparison with more baselines.} 
We also include Consistency Policy~\cite{prasadConsistencyPolicyAccelerated2024} as our baseline. Due to the time-consuming procedure of training a teacher model and then distilling it into a student model in Consistency Policy pipeline, we only evaluate its performance on a few randomly selected tasks across various domains. In RoboMimic tasks, we follow the original settings in its paper. For Push-T and Meta-World tasks, we run its pipeline using our settings.
The results are reported in Table~\ref{table:compare with more baselines}. \ours-Policy shows consistent improvement on all benchmarks.

\textbf{Ablation study on reflow and \textit{\ours}.}
Some works argue that although the distillation of the diffusion process can be used to accelerate policy synthesis, it is computationally expensive and can hurt both the accuracy and diversity of synthesized actions. In this paper, we show that although this statement holds true in some circumstances, we can still neutralize the drawbacks of reflow by our \textit{\ours} method. Here we test the performance of base policy, reflow policy, and \ours-Policy. We choose 3 tasks with significantly different base policy accuracy on Meta-World and Adroit to prove our point. As presented in Table~\ref{table:rf deisgn}, on tasks with low success rate, such as pen and assembly, reflow may lead to worse performance. However, the \ours-Policy retains high score when it comes to base policy 100\% accuracy tasks, such as the plate-slide-side task. It proves our hypothesis that the performance degradation brought by reflow is a kind of error accumulation. Fortunately, our proposed \ours technique compensates for accuracy loss and achieves a much higher score, sometimes even better than base policy. 
Note that $\pi_0$~\cite{pi0} is equivalent to the first policy in Table~\ref{table:rf deisgn}. Although it is also trained via the flow matching loss, $\pi_0$ does not involve any reflow or \textit{\oursfull}, which effectiveness is underlined by the numbers in the last row in Table~\ref{table:rf deisgn}. This means that vanilla flow-based policies can be strengthened with our proposed methods, since ours can be implemented upon $\pi_0$ or any other policy that utilizes flow matching loss. 

\begin{table*}[t]
\centering
\vspace{0.1in}
\caption{\textbf{Main simulation results.} Averaged over 66 tasks, \ours achieves $\mathbf{50.1\%}$ relative improvement compared to Diffusion Policy. The complete results are shown in Table~\ref{table:all simulations}.
}
\label{table:main simulation}
\resizebox{1.0\textwidth}{!}{%
\begin{tabular}{l|cccccccc|cc}
\toprule

\multirow{2}{*}{Algorithm $\backslash$ Task} & Adroit &  RoboMimic & Kitchen & Push-T & Meta-World & Meta-World   & Meta-World   & Meta-World  &  \multicolumn{2}{c}{Average}\\

 & (3) & (5)  & (7)  & (1) & Easy (28) & Medium (11) & Hard (6) & Very Hard (5)& \multicolumn{2}{c}{(66)} \\

\midrule

\textbf{\ours} &  \ccbf{41.3} & \ccbf{95.4} & \ccbf{58.1} & \ccbf{80.0} & \ccbf{78.1} & \ccbf{34.5}  & \ccbf{42.2} & \ccbf{41.4} &  \multicolumn{2}{c}{\ccbf{62.3} ($\uparrow \mathbf{50.1}\%$)}\\

Diffusion Policy & \cc{25.0} & \cc{93.2} & \cc{57.1} & \cc{78.0} & \cc{38.3} & \cc{19.5} & \cc{16.7} & \cc{32.0} & \multicolumn{2}{c}{\cc{38.9}}\\

AdaFlow Policy & \cc{33.7} & \cc{95.0} & \cc{49.2} & \cc{72.0} & \cc{46.2} & \cc{26.7} & \cc{17.8} & \cc{12.0} & \multicolumn{2}{c}{\cc{41.5}}\\

\bottomrule
\end{tabular}}
\vspace{-0.25in}
\end{table*}

\begin{table}[t]
\begin{minipage}[t]{0.48\textwidth}
    \centering
    \captionof{table}{\textbf{Comparing \ours with more baselines in simulation.}}
    \label{table:compare with more baselines}
    \resizebox{\textwidth}{!}
    {%
    \begin{tabular}{l|c|c|c|c}
    \toprule
    
    \multirow{2}{*}{Algorithm $\backslash$  Task}  & {\textbf{Adroit}} & \multirow{2}{*}{\textbf{Push-T}} & \textbf{RoboMimic} & \multirow{2}{*}{\textbf{Average}} \\
      & Pen & & Square &  \\
    
    \midrule
    
    \textbf{\ours} & \ddbf{52}{8} &\ddbf{80}{2}& \ddbf{97}{5} & \ccbf{76.3}\\
    
    Diffusion Policy (100-step) &\dd{20}{7}&\dd{78}{1} & \dd{83}{2} & \cc{60.3}\\
    Diffusion Policy (1-step) & \dd{0}{0} & \dd{70}{0} & \dd{0}{0} & \cc{23.3}\\
    
    Consistency Policy (1-step) &\dd{32}{8}   & \dd{71}{2}& \dd{89}{2}  & \cc{64.0}\\
    
    AdaFlow Policy (20-step) &\dd{37}{9}   & \dd{72}{0} & \dd{90}{3}  & \cc{66.3}\\
    \bottomrule
    \end{tabular}}
\vspace{0.1in}
\end{minipage}
\hfill
\begin{minipage}[t]{0.48\textwidth}
\centering
\captionof{table}{\textbf{Inference latency for one action prediction step in simulation (ms).} }
\label{table:inference latency}
\resizebox{.8\textwidth}{!}
{%
\begin{tabular}{l|c|c}
\toprule

    Algorithm & NFE & Inference Latency (ms)   \\

\midrule

\ours & 1 & 16.72 \\

Diffusion Policy & 100 & 1287 \\

Consistency Policy & 1 & 18.35 \\

AdaFlow Policy & 20 & 629.0 \\

1-step DDIM & 1 & 16.16 \\

\bottomrule
\end{tabular}}
\end{minipage}
\end{table}

\begin{table}[!tbp]
\centering
\vspace{-0.05in}
\caption{\textbf{Ablation on the selective flow alignment.}
}
\label{table:rf deisgn}
\resizebox{\linewidth}{!}
{%
\tabcolsep=0.12cm
\begin{tabular}{l|c|c|cc|c}
\toprule

\multirow{2}{*}{Designs $\backslash$ Task}& \multirow{2}{*}{\#Steps}& \textbf{Adroit} & \multicolumn{2}{c|}{\textbf{Meta-World}} & \multirow{2}{*}{\textbf{Average}}\\
  &  & Pen & Assembly & Plate Slide Side & \\
\midrule

Base Policy & 100 &   \dd{43}{6} & \dd{27}{9} & \ddbf{100}{0} & \cc{56.7} \\

Reflow Policy~\cite{liu2023flow} & 1 &  \dd{40}{8} & \dd{7}{9} & \ddbf{100}{0} & \cc{49.0} \\

\ours & 1 &  \ddbf{52}{8} & \ddbf{33}{9}  & \ddbf{100}{0} & \ccbf{61.7}\\

\bottomrule
\end{tabular}}
\vspace{-0.2in}
\end{table}

\textbf{Ablation study on 1-step solver.} A 1-step Euler solver works well in the default \ours-Policy inference settings. To further explore it effectiveness, we replace it with a 100-step solver. Intuitively, a solver with more steps contributes to a higher accuracy since the truncation error introduced by using an approximation is decreased in every denoising loop. 
We also evaluate the performance of the Runge-Kutta method of order 5(4) from Scipy~\cite{rk45}, denoted as RK45. The number of steps is adaptively decided based on user-specified relative and absolute tolerances, with its minimum no less than 1. 
As shown in Table~\ref{table:solver}, the RK45 solver reports the highest accuracy for reflow policy. For \ours-Policy, the 1-step solver is on a par with the 100-step one. 
Notwithstanding the comparable success rates, a 1-step solver exceeds its 100-step counterpart in inference efficiency by approximately two orders of magnitude. Since the number of sampling steps of RK45 is not fixed to 1, it still lags behind \ours-Policy with respect to efficiency. 

\vspace{0.05in}
\begin{table}[t]
\centering
\vspace{0.1in}
\caption{\textbf{Ablation on solver choices.} We compare the effectiveness of the 1-step Euler solver with a 100-step solver and a RK45 solver with adaptive timesteps.}
\label{table:solver}
\resizebox{0.5\textwidth}{!}
{%
\begin{tabular}{l|l|c|cc|c}
\toprule

\multicolumn{2}{c|}{\multirow{2}{*}{Solver $\backslash$ Task}}& \textbf{Adroit} & \multicolumn{2}{c|}{\textbf{Meta-World}} & \multirow{2}{*}{\textbf{Average}}\\
\multicolumn{2}{c|}{} 
 & Pen & Assembly & Plate Slide Side & \\
\midrule

\multirow{3}{*}{Base Policy} & 1-step solver & \dd{43}{6}  & \dd{27}{9} & \ddbf{100}{0} &  \dd{57}{5} \\
& 100-step solver  & \dd{45}{7} & \ddbf{30}{6} & \ddbf{100}{0} & \dd{58}{4} \\
& RK45 solver  & \ddbf{46}{7} & \dd{30}{9} & \ddbf{100}{0} & \ddbf{59}{5} \\
\midrule

 & 1-step solver  & \dd{40}{8} & \dd{7}{9} & \ddbf{100}{0} & \dd{49}{6} \\
Reflow Policy& 100-step solver   & \dd{40}{4} & \dd{25}{9} & \ddbf{100}{0} & \dd{55}{4} \\
& RK45 solver  & \ddbf{45}{6} & \ddbf{30}{8}  & \ddbf{100}{0} & \ddbf{58}{5} \\
\midrule

 & 1-step solver  & \dd{52}{8} & \dd{33}{9} & \ddbf{100}{0} & \dd{62}{6} \\
\ours-Policy & 100-step solver  & \ddbf{52}{6} & \ddbf{33}{6} & \ddbf{100}{0} & \ddbf{62}{4} \\
& RK45 solver & \dd{50}{4} & \dd{31}{4} &\ddbf{100}{0} & \dd{60}{3} \\

\bottomrule
\end{tabular}}
\end{table}

\subsection{Efficiency}

We evaluate inference latency on an NVIDIA RTX 6000 Ada GPU on the Meta-World assembly task. \ours-Policy requires only a single denoising step per action prediction, while Diffusion Policy requires 100 steps as per original settings and AdaFlow 20. With observation encoding consuming negligible time compared to sampling, \ours-Policy achieves dramatic speedup.

After a 200-iteration warm-up, we measure average latency over 800 rollouts, excluding the upper and lower duration quartiles. Table~\ref{table:inference latency} presents wall clock times for each algorithm. \ours-Policy achieves \textbf{98.7\%} and \textbf{97.3\%} acceleration compared to Diffusion Policy and AdaFlow Policy respectively, while maintaining superior accuracy across all domains. Consistency Policy achieves comparable inference speed with its 1-step sampling approach but requires extra training computation when distilling its teacher model. Consistency Policy underperforms compared to \ours-Policy despite similar inference speed. 1-step DDIM inference shows comparable speed to \ours-Policy but significantly sacrifices performance, failing in nearly all tasks.

\subsection{Robustness}
\begin{figure}[htbp]
    \centering
    \includegraphics[width=0.8\linewidth]{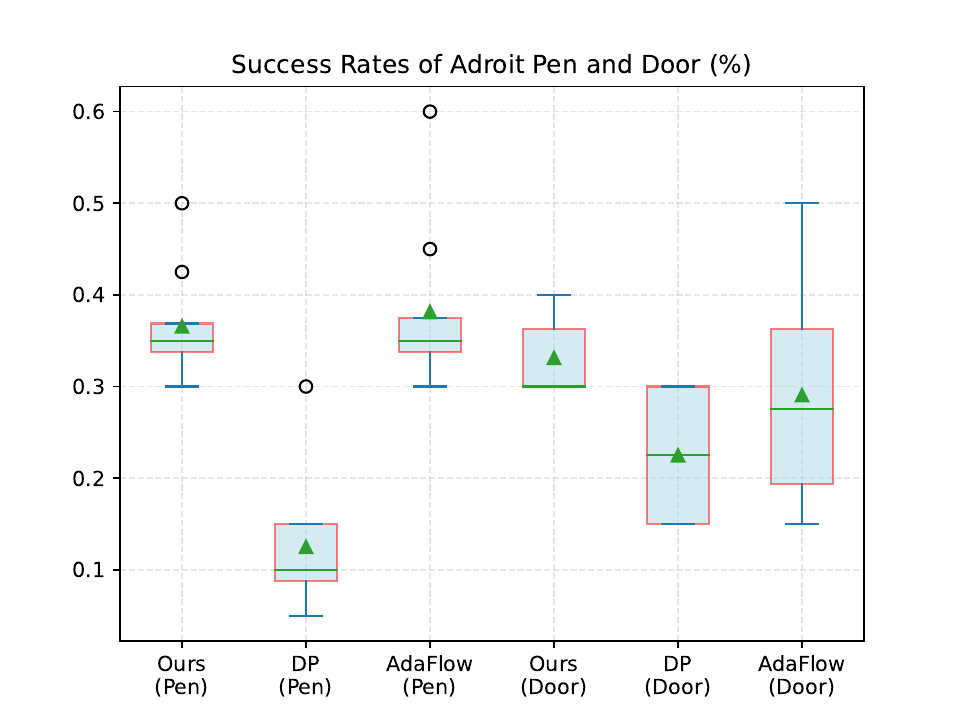}
    \captionof{figure}{\textbf{Success Rates on Adroit (\%).}}
    \label{fig:variance}
\end{figure}

During evaluation, we observed that \ours-Policy shows more robustness than Diffusion Policy and AdaFlow Policy. Take Adroit door and pen as examples. Evaluated on one checkpoint iteratively using different inference seeds, our algorithm sees a lower variance in accuracy than that of baselines. This is an extremely important attribute since in diffusion-based algorithms, different noise latent initializations are heavily related to the final performance. A well-generalized algorithm should maintain a steady success rate among different initial noise samples to avoid fluctuation in performance during random evaluation.
The robustness of \ours-Policy may be mainly attributed to the straight and stable sampling trajectories of rectified flow modeling.  
Results are shown in Figure~\ref{fig:variance}.

\section{Real World Experiments}

To comprehensively evaluate our method in the real world, we select 7 representative manipulation tasks, as shown in Table~\ref{table:real task summary}. 
The tasks cover a variety of motions, including twist, pick, place, push, pull, and sweep, posing challenges in terms of precision, dexterity, and long-horizon.

\subsection{Experimental Analysis}

\begin{table}
\centering
\vspace{0.2cm}
\caption{\textbf{Main results for real robot experiments.} The first 3 and last 4 tasks are evaluated with 10 and 20 trials each, respectively.}
\label{table:real task summary}
\resizebox{0.5\textwidth}{!}{%
\begin{tabular}{lcccclcccc}
\toprule
\multicolumn{3}{c}{\textbf{Real Robot Benchmark (7 Tasks) }} \\
\midrule

Task & Diffusion Policy & \ours \\

\midrule
\textbf{Putting Apple in the Bowl} & \ccbf{100\%} & \ccbf{100\%}   \\
\textbf{Moving Object Picking} & 60\% & \ccbf{70\%}  \\
\textbf{Flower Insertion} & 20\% & \ccbf{80\%}   \\
\textbf{Coffee Bean Sweeping} & 35\% & \ccbf{70\%}  \\
\textbf{Drawer Close} & 40\% & \ccbf{60\%}  \\
\textbf{Rice Pouring} & 0\% & \ccbf{30\%}  \\
\textbf{Knob Pull} & 0\% & \ccbf{40\%}  \\

\bottomrule
\end{tabular}
}
\vspace{-0.15in}
\end{table}

\textbf{Real-world manipulation results.} Table~\ref{table:real task summary} presents a comparative analysis of success rates between \ours-Policy and Diffusion Policy across various real-world robotic manipulation tasks. The empirical evidence demonstrates that \ours-Policy consistently outperforms Diffusion Policy in tasks demanding high precision, dexterity, and long-horizon sequential manipulation. Diffusion Policy exhibits inferior performance primarily because it is inherently unstable, with success rates substantially varying across different noise initializations. In the Putting Apples in the Bowl task, both policies achieve perfect success rates (100\%), indicating comparable capabilities in handling basic pick-and-place operations with irregularly shaped objects. However, the performance gap widens significantly in favor of \ours-Policy as task complexity increases. For the Moving Object Picking, which involves handling a rubber duck on a dynamic water surface, \ours-Policy achieves a 70\% success rate compared to Diffusion Policy's 60\%. This superiority can be attributed to the continuous rectification mechanism in \ours-Policy, which facilitates more robust and adaptive visuomotor control when managing complex interactions between the gripper, floating objects, and water surface. In the most challenging tasks—Flower Insertion and Rice Pouring—which involve long-horizon sequential manipulation consisting of precise grasping and delicate wrist twisting, \ours-Policy achieves 80\% and 30\% success rate, substantially outperforming Diffusion Policy's 20\% and 10\%. This marked improvement underscores the efficacy of \ours-Policy in generating consistent and accurate action sequences for multi-stage, fine-grained continuous manipulation scenarios.

\begin{figure}[!t]
\begin{center}
    \centering
    \captionsetup{type=figure}
\subfigure[\textbf{Apple.}]{
\includegraphics[width=0.23\textwidth]{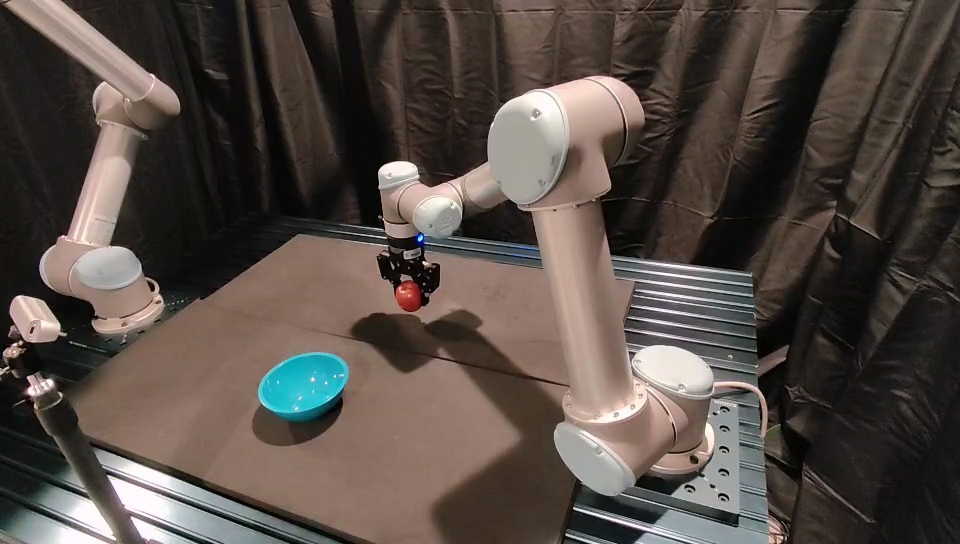}}
\subfigure[\textbf{Orange cube.}]{
\includegraphics[width=0.23\textwidth]{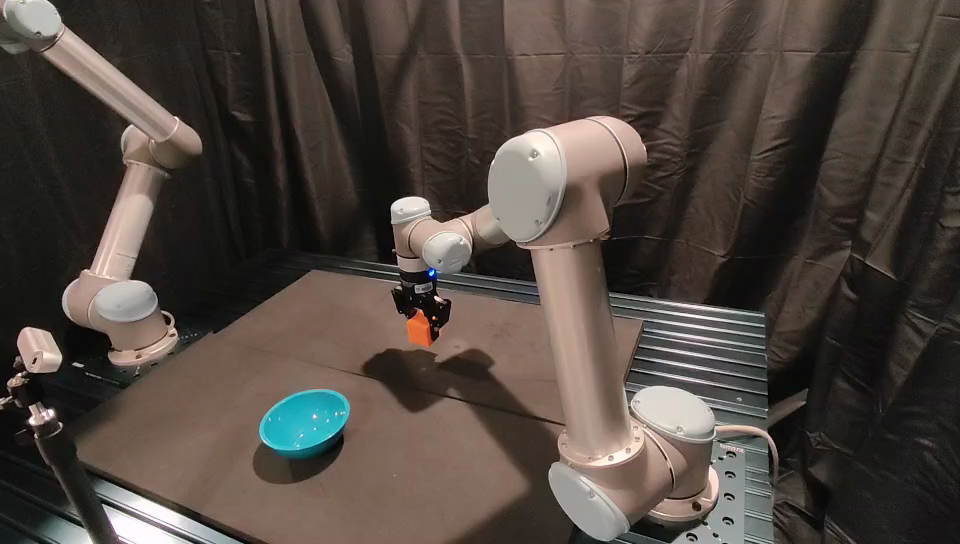}}
\subfigure[\textbf{Rubber duck.}]{
\includegraphics[width=0.23\textwidth]{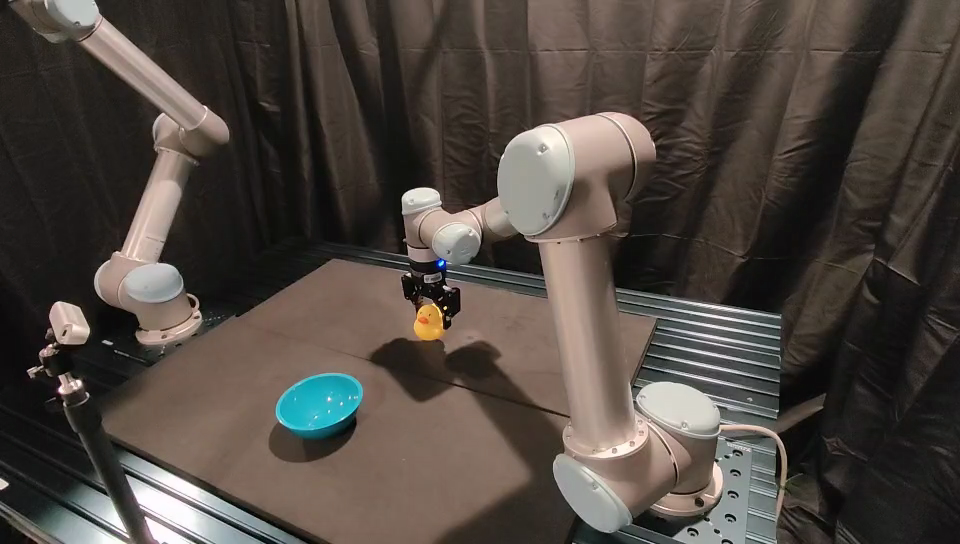}}
\vspace{-0.3cm}
\caption{\textbf{Grasping different objects with one policy.} \ours trained on the apple can generalize to other objects (cube, rubber duck) with similar sizes and locations.}
\vspace{-0.6cm}
\label{fig:real-obj-gen}
\end{center}
\end{figure}

\begin{figure}[!t]
\begin{center}
    \centering
    \captionsetup{type=figure}
\includegraphics[width=0.35\textwidth]{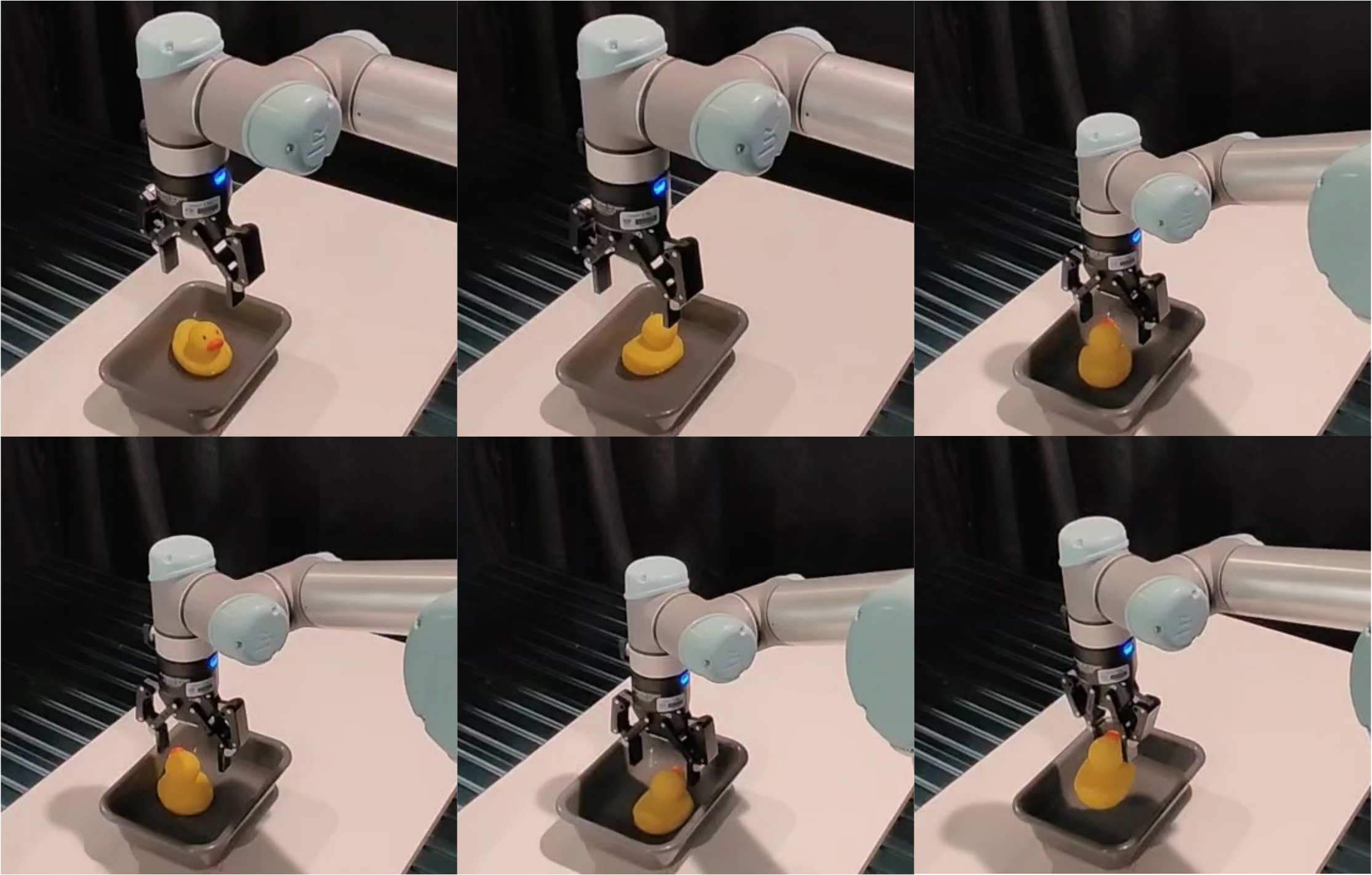}
\caption{\textbf{Floating object manipulation.} \ours dynamically adjusts its action trajectory to approach and grab the moving rubber duck on the water, which demonstrates generalization ability to different object locations.}
\vspace{-0.5cm}
\label{fig:loc-gen}
\end{center}
\end{figure}
\begin{figure}[t]
\vspace{0.2cm}
\begin{center}
    \centering
    \captionsetup{type=figure}
\includegraphics[width=0.4\textwidth]{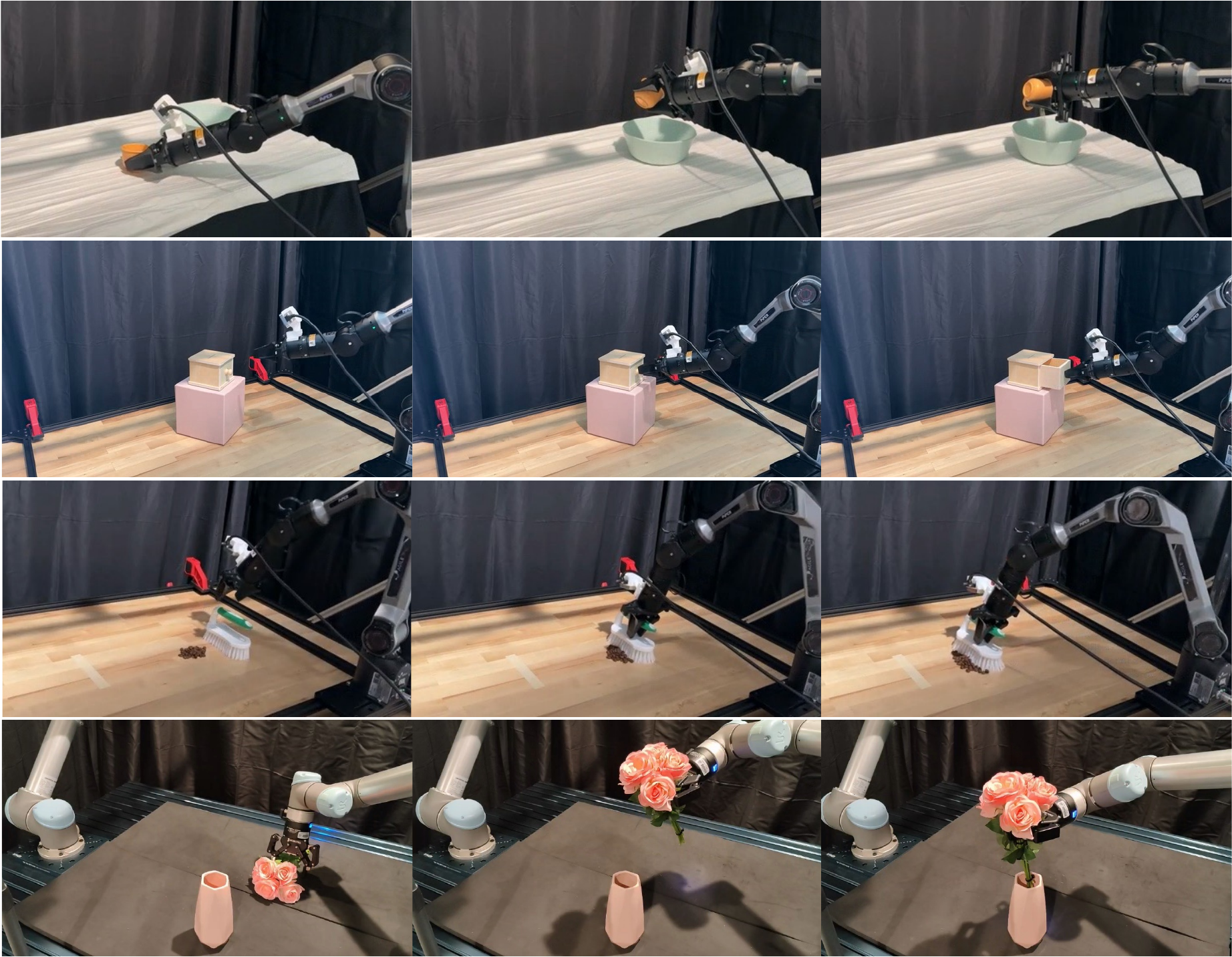}
\vspace{-0.05cm}
\caption{\textbf{Precise manipulation.} \ours successfully performs Rice Pouring, Knob Pull, Coffee Bean Sweeping, and Flower Insertion (from top to bottom).}
\vspace{-0.9cm}
\label{fig:all_task}
\end{center}
\end{figure}

\textbf{Handling different object appearance.} As shown in Figure~\ref{fig:real-obj-gen}, while trained on grasping one object, the policy successfully generalizes to others in various appearances, suggesting that \ours-Policy can leverage learned visuomotor patterns to handle variations in object appearance while maintaining successful grasp execution.

\textbf{Manipulating dynamic objects.} Figure~\ref{fig:loc-gen} illustrates how \ours-Policy adjusts its action trajectory to approach and grasp a moving rubber duck on water. Unlike static object grasping, this task requires the policy to continuously refine its motion based on the ever-changing real-time position. 
The sequential images indicate that \ours-Policy can generate high-frequency visuomotor actions to compensate for object drift, allowing successful execution even towards unfixed location. This highlights its suitability for real-world tasks involving dynamic objects.

\textbf{Precise manipulation.} Flower Insertion and Rice Pouring involve multi-stage manipulation, i.e., first grasping and then performing twists in one or several joints. Rice Pouring is more demanding, which requires an exact wrist twisting whilst other joints remain still to ensure the rice grains fall within the bowl. Knob Pull and Drawer Close require the end effector to perform a pull/push towards a straight line, otherwise the drawer would stuck by friction. Also, Knob Pull includes precise grasping before the smooth pull action. In Coffee Bean Sweeping, the robot arm should apply a force onto the tabletop, which needs precise height control.

\section{Conclusion}
In this work, we presented \ours, visuomotor policy learning framework that supports efficient and accurate inference. \ours is a flow-based model that facilitates single-step action synthesis and addresses the performance degradation by introducing a novel selective flow alignment strategy. Through rigorous real-world and simulation evaluations, \ours exhibited remarkable performance across tasks demanding high precision, dexterity, and extended temporal planning horizons. Comparative analyses demonstrate that \ours substantially outperforms state-of-the-art approaches, achieving superior success rates while concurrently reducing inference latency by 98.7\%, thereby enabling significantly more responsive and stable robot execution. These empirical findings underscore the efficacy of \ours as a computationally efficient and highly effective solution for real-time visuomotor control in sophisticated manipulation tasks, with promising implications for future deployments in dynamic robotic applications.

\textbf{Acknowledgements.}
The USC Physical Super Intelligence Lab acknowledges generous supports from Toyota Research Institute, Dolby, Google DeepMind, Capital One, Nvidia, and Qualcomm. This work was partially supported by the National Science Foundation through NSF CPS \#2434460. Yue Wang is also supported by a Powell Research Award.
\section{Appendix}

\subsection{Experiment Setup in Simulation}~\label{sec:exp setup}
In this section, we provide an overview of our simulation domains, our evaluation methodology on the tasks, and our core findings.

We use DDIM~\cite{song2021denoising} as the noise scheduler and predict the vector $\mathbf{a}_0 - \mathbf{a}_T$ instead of epsilon or sample prediction, with 100 timesteps during training. 
We train 1000 epochs for Meta-World and Adroit tasks given their simplicity and 3050 epochs for RoboMimic tasks and Push-T. For Franka Kitchen, we train 5000 epochs due to its long-horizon and multi-task complexity.
Real-world tasks are trained with 1000 epochs. 
The optimizer used is AdamW with the same hyperparameters as that used in ~\cite{chi2023diffusion}.
Batch size is 64 for \ours Policy and all the baselines except Franka Kitchen where the batch size is 256.

Before the \ours phase, we sample 10 couplings in each action space, saving both generation time and \ours Policy training time. 

\subsection{Main Results}

Here we present the main results of all experiments in simulation in Table~\ref{table:all simulations}. 

\begin{table*}[htbp]
\centering
\caption{\textbf{Main results on 66 simulation tasks.} Results for all tasks are provided in this table.
A summary across domains is shown in Table~\ref{table:main simulation}.
}
\label{table:all simulations}
\resizebox{0.8\textwidth}{!}{%
\begin{tabular}{l|ccc|ccccc}
\toprule
& \multicolumn{3}{c|}{\textbf{Adroit}~\cite{rajeswaran2017dapg}} & \multicolumn{5}{c}{\textbf{RoboMimic}~\cite{mandlekar2021matters}} \\

 Alg. $\backslash$ Task & Pen  & Door & Hammer & Lift & Can & Square & Transport & Tool Hang \\
 
\midrule
 
\ours & \ddbf{52}{8} & \ddbf{37}{5} & \dd{35}{4} & \ddbf{100}{0} & \ddbf{100}{0} & \dd{91}{3} &  \ddbf{89}{5} & \ddbf{97}{5} \\
 Diffusion Policy & \dd{20}{7} & \dd{23}{6}  & \dd{32}{6} & \ddbf{100}{0} & \dd{99}{1}  & \ddbf{95}{3} & \dd{89}{7} & \dd{83}{2} \\
 AdaFlow Policy & \dd{37}{9} & \dd{28}{9} &  \ddbf{36}{6} & \ddbf{100}{0} & \ddbf{100}{0} & \dd{90}{0} & \dd{76}{6} & \dd{60}{3}  \\
\bottomrule
\end{tabular}}

\resizebox{0.8\textwidth}{!}{%
\begin{tabular}{l|ccccccc|c}
\toprule
&  \multicolumn{7}{c|}{\textbf{Kitchen}~\cite{lynch2019play}} & \multirow{2}{*}{\textbf{Push-T}~\cite{florence2022ibc}}\\

 Alg. $\backslash$ Task & p1 & p2 & p3 & p4 
 & p5 & p6 & p7 
 &\\
 
\midrule

\ours &  \ddbf{100}{0}  & \ddbf{100}{0}  & \ddbf{100}{0}  & \ddbf{100}{0} & \ddbf{7}{3}  & \ddbf{0}{0} & \ddbf{0}{0} & \ddbf{80}{2} \\
 Diffusion Policy & \ddbf{100}{0}  & \ddbf{100}{0} & \dd{99}{1} & \dd{98}{2}  &  \dd{3}{2} & \ddbf{0}{0} & \ddbf{0}{0}  & \dd{78}{1} \\
 AdaFlow Policy & \dd{99}{1} & \dd{99}{1} & \dd{82}{3} & \dd{63}{5} & \dd{1}{2} & \ddbf{0}{0} & \ddbf{0}{0} & \dd{72}{0} \\
\bottomrule
\end{tabular}}

\resizebox{\textwidth}{!}{%
\begin{tabular}{l|cccccccc}
\toprule
& \multicolumn{7}{c}{\textbf{Meta-World~\cite{yu2020metaworld} (Easy)}} \\

 Alg. $\backslash$ Task & Button Press & Button Press Topdown & Button Press Topdown Wall & Button Press Wall & Coffee Button & Dial Turn & Door Close \\
\midrule

\ours & \dd{87}{9} & \ddbf{87}{9} & \ddbf{93}{9} & \dd{93}{9} & \ddbf{100}{0} & \ddbf{67}{9} &  \ddbf{100}{0}  \\
 Diffusion Policy & \dd{67}{9} & \dd{67}{9} & \dd{67}{19} & \ddbf{100}{0} & \dd{93}{9} & \dd{27}{9} &  \dd{0}{0}  \\
 AdaFlow Policy & \ddbf{93}{9} & \dd{53}{9} & \dd{87}{9} & \dd{87}{9} & \ddbf{100}{0} & \dd{27}{9} &  \dd{87}{9}  \\

\bottomrule
\end{tabular}}

\resizebox{\textwidth}{!}{%
\begin{tabular}{l|ccccccccccc}
\toprule
& \multicolumn{9}{c}{\textbf{Meta-World (Easy)}} \\

 Alg. $\backslash$ Task & Door Lock & Door Open & Door Unlock & Drawer Close & Drawer Open & Faucet Close & Faucet Open & Handle Press & Handle Pull\\
\midrule
\ours & \ddbf{27}{9} & \ddbf{93}{9} & \ddbf{100}{0} & \ddbf{100}{0} & \ddbf{100}{0} & \dd{87}{9} &  \ddbf{93}{9} & \dd{73}{25} &  \ddbf{13}{9}    \\
 Diffusion Policy &  \dd{20}{16} & \dd{20}{0} & \dd{93}{9} & \dd{67}{25} & \dd{93}{9} & \ddbf{93}{9} &  \dd{53}{9} & \dd{47}{9} &  \dd{6}{9}  \\
 AdaFlow Policy & \dd{20}{0} & \dd{47}{9} & \dd{87}{9}  & \ddbf{100}{0} & \dd{93}{9} &  \ddbf{93}{9} & \dd{87}{9} & \ddbf{93}{9} &  \dd{0}{0} \\
 
\bottomrule
\end{tabular}}

\resizebox{\textwidth}{!}{%
\begin{tabular}{l|ccccccccccc}
\toprule
& \multicolumn{8}{c}{\textbf{Meta-World (Easy)}} \\

 Alg. $\backslash$ Task & Handle Press Side & Handle Pull Side & Lever Pull & Plate Slide & Plate Slide Back & Plate Slide Back Side & Plate Slide Side & Reach\\
\midrule

\ours &  \ddbf{93}{9} & \ddbf{27}{9} & \ddbf{60}{16} & \ddbf{87}{9} & \dd{73}{9} & \ddbf{100}{0} &  \ddbf{100}{0} & \dd{33}{9} \\

Diffusion Policy & \dd{20}{16} & \dd{0}{0} & \dd{20}{0} & \dd{80}{0} & \dd{73}{19} & \dd{67}{25} &  \ddbf{100}{0} & \ddbf{40}{0} \\
 AdaFlow Policy & \dd{87}{9} & \dd{13}{9} & \dd{33}{9} & \dd{40}{0} & \ddbf{80}{0} & \ddbf{100}{0} & \dd{80}{0} & \dd{7}{9} \\

\bottomrule
\end{tabular}}

\resizebox{\textwidth}{!}{%
\begin{tabular}{l|cccc|ccccccc}
\toprule
& \multicolumn{4}{c|}{\textbf{Meta-World (Easy)}} & \multicolumn{5}{c}{\textbf{Meta-World (Medium)}} \\

 Alg. $\backslash$ Task & Reach Wall & Window Close & Window Open & Peg Unplug Side & Basketball & Bin Picking & Box Close & Coffee Pull & Coffee Push\\
\midrule

\ours &  \dd{67}{9} & \ddbf{100}{0} & \ddbf{87}{9} & \ddbf{47}{25} & \ddbf{0}{0} & \ddbf{53}{25} &  \ddbf{27}{19} & \dd{80}{16} & \ddbf{100}{0} \\
 Diffusion Policy & \dd{47}{9} & \dd{73}{19} & \dd{33}{25} & \dd{20}{16} & \ddbf{0}{0} & \dd{0}{0} &  \dd{13}{9} & \dd{27}{9} & \dd{20}{16} \\
 AdaFlow Policy & \ddbf{80}{0} & \ddbf{100}{0} & \dd{33}{9} & \dd{20}{0} & \dd{0}{0} & \dd{27}{9} &  \dd{20}{0} & \ddbf{100}{0} & \dd{87}{9} \\

\bottomrule
\end{tabular}}

\resizebox{\textwidth}{!}{%
\begin{tabular}{l|cccccc|ccccc}
\toprule
& \multicolumn{6}{c|}{\textbf{Meta-World (Medium)}} & \multicolumn{4}{c}{\textbf{Meta-World (Hard)}} \\

 Alg. $\backslash$ Task & Hammer & Peg Insert Side & Push Wall & Soccer & Sweep & Sweep Into & Assembly & Hand Insert & Pick Out of Hole & Pick Place\\
\midrule

\ours & \dd{33}{9} & \ddbf{13}{19} & \dd{13}{19} & \ddbf{27}{9} & \dd{13}{9} & \dd{20}{16} &  \dd{33}{9} & \ddbf{40}{16} & \ddbf{47}{25} & \ddbf{33}{9} \\
Diffusion Policy & \ddbf{67}{19} & \dd{0}{0} & \dd{13}{9} & \dd{20}{16} & \ddbf{27}{9} & \ddbf{27}{19}  & \dd{27}{9} & \dd{27}{9} & \dd{0}{0} & \dd{0}{0} \\
AdaFlow Policy & \dd{20}{0} & \dd{0}{0} & \ddbf{20}{0} & \dd{7}{9} & \dd{0}{0} & \dd{13}{19}  & \ddbf{40}{16} & \dd{7}{9} & \dd{0}{0} & \dd{0}{0} \\

\bottomrule
\end{tabular}}

\resizebox{\textwidth}{!}{%
\begin{tabular}{l|cc|ccccc|ccccc|c}
\toprule
& \multicolumn{2}{c|}{\textbf{Meta-World (Hard)}} & \multicolumn{5}{c|}{\textbf{Meta-World (Very Hard)}} & 
 \multirow{2}{*}{\textbf{Average}} \\

 Alg. $\backslash$ Task & Push & Push Back & Shelf Place & Disassemble & Stick Pull & Stick Push & Pick Place Wall &  \\
\midrule

\ours & \ddbf{47}{9} & \ddbf{53}{19} & \ddbf{7}{9} & \ddbf{87}{9} & \ddbf{13}{19} & \ddbf{80}{16} &  \ddbf{20}{0} & \ccbf{62.2} \\

 Diffusion Policy & \dd{13}{9} & \dd{33}{9} & \dd{0}{0} & \dd{60}{16} & \dd{7}{9} & \ddbf{80}{16} &  \dd{13}{19} & \cc{38.9}\\
 AdaFlow Policy & \dd{40}{16} & \dd{20}{16} & \dd{0}{0} & \dd{53}{9} & \dd{7}{9} &  \dd{0}{0} & \dd{0}{0} &  \cc{40.9} \\

\bottomrule
\end{tabular}}

\end{table*}

\subsection{Ablation}
\subsubsection{3D Inputs}

\begin{figure}[htbp]
\centering
\includegraphics[width=.45\textwidth]{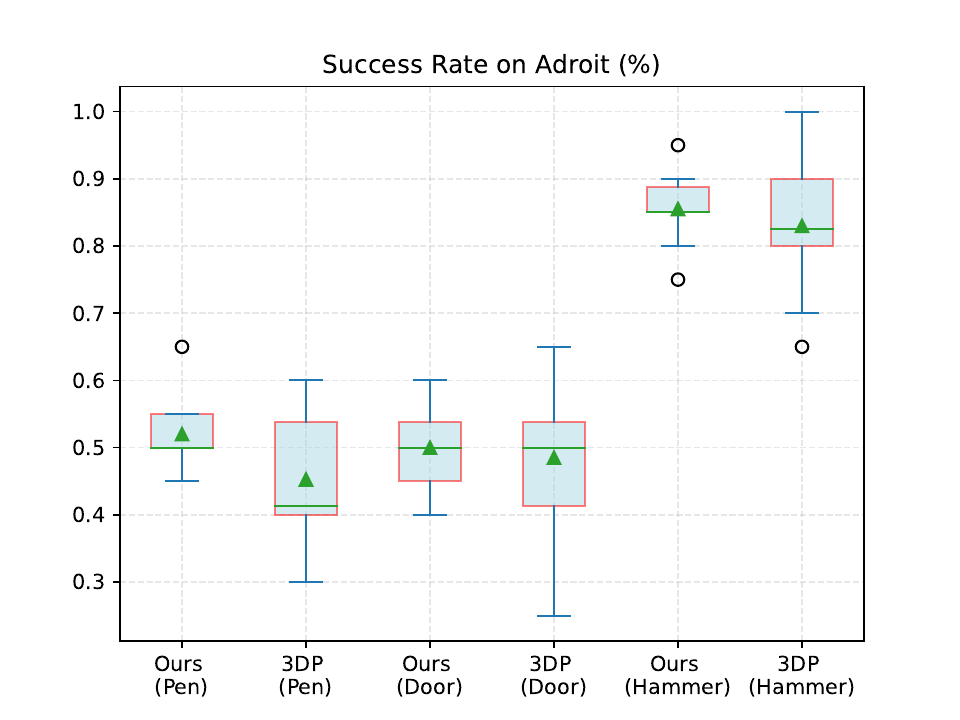}
\captionof{figure}{\textbf{Success Rate on 3D Adroit inputs (\%).} We evaluate 3 simulated tasks to show the accuracy variance of \ours and 3D Diffusion Policy (3DP).}
\label{fig:3d variance}
\end{figure}

To demonstrate our method has the capability to generalize to inputs with other modalities, we take the 3D point clouds as inputs. We compare \ours with 3D Diffusion Policy~\cite{Ze2024DP3} (DP3) on Adroit. We only substitute the policy network for DP3 policy. All the other settings are the same as the baseline, to make a fair comparison. The number of demos is 10, the same as reported in the 3D Diffusion Policy paper. We train the models with seed $0$. We evaluate them with $3$ seeds, and then report the average maximum success rate among them. As shown in Table~\ref{table:3D input}, \ours performs slightly better than 3D Diffusion Policy. Although 3D Diffusion Policy shows similar average accuracy, it is unstable and the performance varies among different initializations of the noise latent. This is reflected in the higher variance of the success rates. We further evaluate the models using $10$ seeds and report the accuracy distribution in Figure~\ref{fig:3d variance}.

\begin{table}[htbp]
\centering
\vspace{-0.1in}
\caption{\textbf{Results on 3D inputs.} We compare \ours and 3D Diffusion Policy on 3D inputs from Adroit.}
\label{table:3D input}
\resizebox{0.45\textwidth}{!}
{%
\begin{tabular}{l|ccc|c}
\toprule

\multirow{2}{*}{Algorithm $\backslash$ Task} &  \multicolumn{3}{c|}{\textbf{Adroit}}  & \multirow{2}{*}{\textbf{Average}} \\

 & Hammer & Door& Pen & \\

\midrule

\ours & \ddbf{95}{4} & \ddbf{57}{6} & \ddbf{59}{4} & \ddbf{70}{5}\\

3D Diffusion Policy & \dd{83}{12} & \dd{53}{12} & \dd{50}{15}& \dd{62}{13}\\

\bottomrule
\end{tabular}}
\end{table}

\bibliographystyle{IEEEtran}
\bibliography{references}

\end{document}